\documentclass[11pt,a4paper]{article}
\usepackage[hyperref]{emnlp-2019}
\usepackage{times}
\usepackage{latexsym}

\usepackage{url}

\usepackage{graphicx}
\usepackage{amsmath}
\usepackage{amssymb}
\usepackage{footnote}
\usepackage{xcolor}
\usepackage{colortbl}
\definecolor{Cyan}{RGB}{0,139,139}
\definecolor{myPurple}{RGB}{160,32,240}
\definecolor{Orange}{cmyk}{.15,.45,1,0}
\definecolor{mygray}{gray}{0.87}
\definecolor{mypink}{RGB}{255,181,197}
\definecolor{cyan}{RGB}{193,227,244}
\definecolor{DeepPink}{RGB}{255,20,147}

\newtheorem{property}{Property}
\newtheorem{proposition}{Proposition}

\newtheorem{hyp}{Assumption} 

\aclfinalcopy 

\title{Learning with Noisy Labels for Sentence-level Sentiment Classification}

\author{Hao Wang~$^{\natural,\, \ddag}$\quad Bing Liu~$^{\ddag,\,}\thanks{\ \ Corresponding author}$ \quad Chaozhuo Li~$^{\S}$\quad Yan Yang~$^{\natural}$\quad Tianrui Li~$^{\natural}$ \\
  $^{\natural}$School of Information Science and Technology, Southwest Jiaotong University \\
  {\tt hwang@my.swjtu.edu.cn, \{yyang;\,trli\}@swjtu.edu.cn} \\
  $^{\ddag}$Department of Computer Science, University of Illinois at Chicago \\
  {\tt liub@uic.edu} \\
  $^{\S}$State Key Lab of Software Development Environment, Beihang University \\
  {\tt lichaozhuo@buaa.edu.cn} \\}

\date{}

\begin{document}
\maketitle
\begin{abstract}
    Deep neural networks (DNNs) can fit (or even over-fit) the training data very well. If a DNN model is trained using data with noisy labels and tested on data with clean labels, the model may perform poorly. This paper studies the problem of learning with noisy labels for sentence-level sentiment classification. We propose a novel DNN model called \textsc{NetAb} (as shorthand for convolutional neural \textsc{\textbf{Net}}works with \textsc{\textbf{Ab}}-networks) to handle noisy labels during training. \textsc{NetAb} consists of two convolutional neural networks, one with a noise transition layer for dealing with the input noisy labels and the other for predicting `clean' labels. We train the two networks using their respective loss functions in a mutual reinforcement manner. Experimental results demonstrate the effectiveness of the proposed model.
\end{abstract}

\section{Introduction}\label{sec_intro}
It is well known that sentiment annotation or labeling is subjective~\cite{liu2012sentiment}. Annotators often have many disagreements. This is especially so for crowd-workers who are not well trained. That is why one always feels that there are many errors in an annotated dataset. In this paper, we study whether it is possible to build accurate sentiment classifiers even with noisy-labeled training data. Sentiment classification aims to classify a piece of text
according to the polarity of the sentiment expressed in the text, e.g., \textit{positive} or \textit{negative} \cite{pang2008opinion,liu2012sentiment,zhang2018deep}. In this work, we focus on sentence-level sentiment classification (SSC) with labeling errors. 

As we will see in the experiment section, noisy labels in the training data can be highly damaging, especially for DNNs because they easily fit the training data and memorize their labels even when training data are corrupted with noisy labels \cite{ZhangBHRV17}. Collecting datasets annotated with clean labels is costly and time-consuming as DNN based models usually require a large number of training examples. Researchers and practitioners typically have to resort to crowdsourcing. However, as mentioned above, the crowdsourced annotations can be quite noisy. 

Research on learning with noisy labels dates back to 1980s \cite{angluin1988learning}. It is still vibrant today \cite{mnih2012learning,natarajan2013learning,natarajan2018cost,menon2015learning,gao2016risk,liu2016classification,khetan2018learning,ZhanWRL19} as it is highly challenging. We will discuss the related work in the next section.

This paper studies the problem of learning with noisy labels for SSC. Formally, we study the following problem.

\textit{Problem Definition}: Given noisy labeled training sentences $S=\{(x_1,y_1),...,(x_n,y_n)\}$, where $x_i|_{i=1}^n$ is the $i$-th sentence and $y_i\in\{1,...,c\}$ is the sentiment label of this sentence, the noisy labeled sentences are used to train a DNN model for a SSC task. The trained model is then used to classify sentences with clean labels to one of the $c$ sentiment labels.



\begin{figure*}[!htb]
    \centering
    \includegraphics[width=1.0\textwidth]{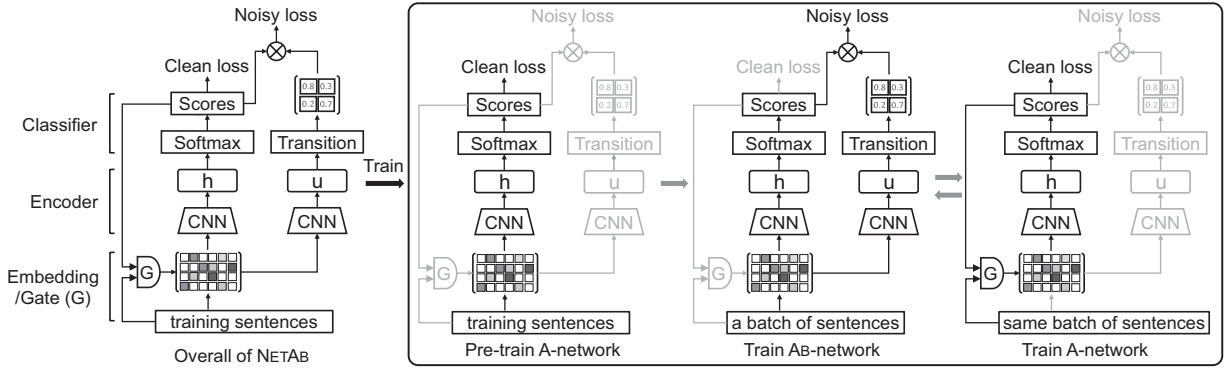}
    \caption{The proposed \textsc{NetAb} model (left) and its training method (right). Components in light gray color denote that these components are deactivated during training in that stage. (Color online)}
    \label{framework}
\end{figure*}

In this paper, we propose a convolutional neural \textsc{\textbf{Net}}work with \textsc{\textbf{Ab}}-networks (\textsc{NetAb}) to deal with noisy labels during training, as shown in Figure \ref{framework}. We will introduce the details in the subsequent sections. Basically, \textsc{NetAb} consists of two  convolutional neural networks (CNNs) (see Figure \ref{framework}), one for learning sentiment scores to predict `clean'~\footnote{Here we use clean with single quotes as it is not completely clean. In practice, models can hardly produce completely clean labels.} labels and the other for learning a noise transition matrix to handle input noisy labels. We call the two CNNs \textsc{A}-network and \textsc{Ab}-network, respectively. The fundamental here is that (1) DNNs memorize easy instances first and gradually adapt to hard instances as training epochs increase \cite{ZhangBHRV17,arpit2017closer}; and (2) noisy labels are theoretically flipped from the clean/true labels by a noise transition matrix \cite{sukhbaatar2015training,goldberger2017training,han2018masking,han2018co}. We motivate and propose a CNN model with a transition layer to estimate the noise transition matrix for the input noisy labels, while exploiting another CNN to predict `clean' labels for the input training (and test) sentences. In training, we pre-train \textsc{A}-network in early epochs and then train \textsc{Ab}-network and \textsc{A}-network with their own loss functions in an alternating manner. To our knowledge, this is the first work that addresses the noisy label problem in sentence-level sentiment analysis. Our experimental results show that the proposed model outperforms the state-of-the-art methods.

\section{Related Work}\label{sec_relatedWork}
Our work is related to sentence sentiment classification (SSC). SSC has been studied extensively \cite{hu2004mining,pang2005seeing,zhao2008adding,Ramanathan2009,tackstrom2011semi,wang2012baselines,yang2014context,kim2014convolutional,tang2015joint,wu2017sentence,wang2018sentiment}. None of them can handle noisy labels. Since many social media datasets are noisy, researchers have tried to build robust models \cite{gamon2004sentiment,barbosa2010robust,liu2012emoticon}. However, they treat noisy data as additional information and don't specifically handle noisy labels. A noise-aware classification model in \cite{ZhanWRL19} trains using data annotated with multiple labels. \citet{wang2016sentiment} exploited the connection of users and noisy labels of sentiments in social networks. Since the two works use multiple-labeled data or users' information (we only use single-labeled data, and we do not use any additional information), they have different settings than ours.

Our work is closely related to DNNs based approaches to learning with noisy labels. DNNs based approaches explored three main directions: (1) training DNNs on selected samples \cite{malach2017decoupling,jiang2018mentornet,ren2018learning,han2018co}, (2) modifying the loss function of DNNs with regularization biases \cite{mnih2012learning,jindal2016learning,patrini2017making,ghosh2017robust,ma2018dimensionality,zhang2018generalized}, and (3) plugging an extra layer into DNNs \cite{sukhbaatar2015training,bekker2016training,goldberger2017training,han2018masking}. All these approaches were proposed for image classification where training images were corrupted with noisy labels. Some of them require noise rate to be known a priori in order to tune their models during training \cite{patrini2017making,han2018co}. Our approach combines direction (1) and direction (3), and trains two networks jointly without knowing the noise rate. We have used five latest existing methods in our experiments for SSC. The experimental results show that they are inferior to our proposed method.  

In addition, \citet{xiao2015learning}, \citet{reed2014training}, \citet{guan2016weakly}, \citet{li2017learning}, \citet{veit2017learning}, and \citet{vahdat2017toward} studied weakly-supervised DNNs or semi-supervised DNNs. But they still need some clean-labeled training data. We use no clean-labeled data.

\section{Proposed Model}\label{sec_model}
Our model builds on CNN \cite{kim2014convolutional}. The key idea is to train two CNNs alternately, one for addressing the input
noisy labels and the other for predicting `clean' labels. The overall architecture of the proposed model is given in Figure \ref{framework}. Before going further, we first introduce a proposition, a property, and an assumption below.
\begin{proposition}\label{prop}
Noisy labels are 
flipped from clean labels by an unknown noise transition matrix.
\end{proposition}

Proposition \ref{prop} is reformulated from~\cite{han2018masking} and has been investigated in \cite{sukhbaatar2015training,goldberger2017training,bekker2016training}. This proposition shows that if we know the noise transition matrix, we can use it to recover the clean labels. In other words, we can put noise transition matrix on clean labels to deal with noisy labels. Given these, we ask the following question: \textit{How to estimate such an unknown noise transition matrix}?

Below we give a solution to this question based on the following property of DNNs.
\begin{property}\label{property}
DNNs tend to prioritize memorization of simple instances first and then gradually memorize hard instances~\cite{ZhangBHRV17}. 
\end{property}

\citet{arpit2017closer} further investigated this property of DNNs.
Our setting is that simple instances are sentences of clean labels and hard instances are those with noisy labels. We also have the following assumption.
\begin{hyp}\label{hyp}
The noise rate of the training data is less than $50\%$.
\end{hyp}

This assumption is usually satisfied in practice because without it, it is hard to tackle the input noisy labels during training.

Based on the above preliminaries, we need to estimate the noisy transition matrix $Q\in \mathbb{R}^{c\times c}$ ($c=2$ in our case, i.e., \textit{positive} and \textit{negative}), and train two classifiers $\ddot{y}\sim P(\ddot{y}|x,\theta)$ and $\widehat{y}\sim\ P(\widehat{y}|x,\vartheta)$, where $x$ is an input sentence, $\ddot{y}$ is its noisy label, $\widehat{y}$ is its `clean' label, $\theta$ and $\vartheta$ are the parameters of two classifiers. Note that both $\ddot{y}$ and $\widehat{y}$ here are the prediction results from our model, not the input labels. We propose to formulate the probability of the sentence $x$ labeled as $j$ with
\begin{equation}
\label{eq1}
    P(\ddot{y}\!=\!j|x,\theta)\!=\!\sum_i\!P(\ddot{y}\!=\!j|\widehat{y}\!=\!i)P(\widehat{y}\!=\!i|x,\vartheta)
\end{equation}
where $P(\ddot{y}=j|\widehat{y}=i)$ is an item (the $ji$-th item) in the noisy transition matrix $Q$. We can see that the noisy transition matrix $Q$ is exploited on the `clean' scores $P(\widehat{y}|x,\vartheta)$ to tackle noisy labels.

We now present our model \textsc{NetAb} and introduce how \textsc{NetAb} performs Eq.~\eqref{eq1}. As shown in Figure \ref{framework}, \textsc{NetAb} consists of two CNNs. The intuition here is that we use one CNN to perform $P(\widehat{y}=i|x,\vartheta)$ and use another CNN to perform $P(\ddot{y}=j|x,\theta)$. Meanwhile, the CNN performing $P(\ddot{y}=j|x,\theta)$ estimates
the noise transition matrix $Q$ to deal with noisy labels. Thus we add a transition layer into this CNN. 

More precisely, in Figure \ref{framework}, the CNN with a clean loss performs $P(\widehat{y}=i|x,\vartheta)$. We call this CNN the \textsc{A}-network. The other CNN with a noisy loss performs $P(\ddot{y}=j|x,\theta)$. We call this CNN the \textsc{Ab}-network. \textsc{Ab}-network shares all the parameters of \textsc{A}-network except the parameters from the Gate unit and the clean loss. In addition, \textsc{Ab}-network has a transition layer to estimate the noisy transition matrix $Q$. In such a way, \textsc{A}-network predict `clean' labels, and \textsc{Ab}-network handles the input noisy labels.

We use cross-entropy with the predicted labels $\ddot{y}$ and the input labels $y$ (given in the dataset) to compute the noisy loss, formulated as below
\begin{equation}
\label{eq2}
    \mathcal{L}_{noisy}\!=\!-\frac{1}{|\ddot{S}|}\sum_{x\in \ddot{S}}\sum_i \mathbb{I}(y\!=\!i|x) \log P(\ddot{y}\!=\!i|x)
\end{equation}
where $\mathbb{I}$ is the indicator function (if $y\!==\!i$, $\mathbb{I}\!=\!1$; otherwise, $\mathbb{I}\!=\!0$), and $|\ddot{S}|$ is the number of sentences to train \textsc{Ab}-network in each batch.

Similarly, we use cross-entropy with the predicted labels $\widehat{y}$ and the input labels $y$ to compute the clean loss, formulated as
\begin{equation}
\label{eq3}
    \mathcal{L}_{clean}\!=\!-\frac{1}{|\widehat{S}|}\sum_{x\in \widehat{S}}\sum_i \mathbb{I}(y\!=\!i|x) \log P(\widehat{y}\!=\!i|x)
\end{equation}
where $|\widehat{S}|$ is the number of sentences to train \textsc{A}-network in each batch.


\begin{table*}[!htb]
    \centering
    \begin{tabular}{|r|c|c|c|c|c|}
    \hline
    & \#Noisy Training Data & \#Clean Training Data & \#Validation Data & \#Test Data \\
    \hline
    Movie & 13539P, 13350N & 4265P, 4265N & 105P, 106N & 960P, 957N \\
    Laptop & 9702P, 7876N & 1064P, 490N & 33P, 20N & 298P, 175N \\
    Restaurant & 8094P, 10299N & 1087P, 823N & 39P, 14N & 339P, 116N \\
    \hline
    \end{tabular} 
    \vspace{-0.05cm}
    \caption{ Summary statistics of the datasets. Number of positive (P) and negative (N) sentences in (noisy and clean) training data, validation data, and test data. The second column shows the statistics of sentences extracted from the 2,000 reviews of each dataset. The last three columns show the statistics of the sentences in three clean-labeled datasets, see ``Clean-labeled Datasets''.}
    \label{table1}
    \vspace{-0.05cm}
\end{table*}

Next we introduce how our model learns the parameters ($\vartheta$, $\theta$ and $Q$). An embedding matrix $v$ is produced for each sentence $x$ by looking up a pre-trained word embedding database (e.g., GloVe.840B~\cite{pennington2014glove}). Then an encoding vector $h\!=\!CNN(v)$ (and $u\!=\!CNN(v)$) is produced for each embedding matrix $v$ in \textsc{A}-network (and \textsc{Ab}-network). A sofmax classifier gives us $P(\hat{y}\!=\!i|x,\vartheta)$ (i.e., `clean' sentiment scores) on the learned encoding vector $h$. As the noise transition matrix $Q$ indicates the transition values from clean labels to noisy labels, we compute $Q$ as follows
\begin{align}
Q &=[q_1;q_2] \\
q_{i} &= softmax(g_if_i),\ i=1,2 \\
g_i &= tanh(W_iu + b_i)
\end{align}
where $W_i$ is a trainable parameter matrix, $b_i$ and $f_i$ are two trainable parameter vectors. They are trained in the \textsc{Ab}-network. Finally, $P(\ddot{y}=j|x,\theta)$ is computed by Eq.~\eqref{eq1}.

In training, \textsc{NetAb} is trained end-to-end. Based on Proposition \ref{prop} and Property \ref{property}, we pre-train \textsc{A}-network in early epochs (e.g., 5 epochs). Then we train \textsc{Ab}-network and \textsc{A}-network in an alternating manner. The two networks are trained using their respective cross-entropy loss. Given a batch of sentences, we first train \textsc{Ab}-network. Then we use the  scores predicted from \textsc{A}-network to select some possibly clean sentences from this batch and train \textsc{A}-network on the selected sentences. Specifically speaking, we use the predicted scores to compute sentiment labels by $\arg\max_i \{\ddot{y}=i|\ddot{y}\sim P(\ddot{y}|x,\theta)\}$. Then we select the sentences whose resulting sentiment label equals to the input label. The selection process is marked by a Gate unit in Figure \ref{framework}. When testing a sentence, we use \textsc{A}-network to produce the final classification result.

\section{Experiments}\label{sec_experiments}
In this section, we evaluate the performance of the proposed \textsc{NetAb} model. we conduct two types of experiments. (1) We corrupt clean-labeled datasets to produce noisy-labeled datasets to show the impact of noises on sentiment classification accuracy. (2) We collect some real noisy data and use them to train models to evaluate the performance of \textsc{NetAb}. 

\textbf{Clean-labeled Datasets.} We use three clean labeled datasets. The first one is the movie sentence polarity dataset from \cite{pang2005seeing}. 
The other two datasets are laptop and restaurant datasets collected from SemEval-2016~\footnote{\url{http://alt.qcri.org/semeval2016/task5/}}. The former consists of laptop review sentences and the latter consists of restaurant review sentences. The original datasets (i.e., Laptop and Restaurant) were annotated with aspect polarity in each sentence. We used all sentences with only one polarity (\textit{positive} or \textit{negative}) for their aspects. That is, we only used sentences with aspects having the same sentiment label in each sentence. Thus, the sentiment of each aspect gives the ground-truth as the sentiments of all aspects are the same.

For each clean-labeled dataset, the sentences are randomly partitioned into training set and test set with $80\%$ and $20\%$, respectively. Following \cite{kim2014convolutional}, We also randomly select $10\%$ of the test data for validation to check the model during training. Summary statistics of the training, validation, and test data are shown in Table \ref{table1}.

\textbf{Noisy-labeled Training Datasets.} For the above three domains (movie, laptop, and restaurant), we collected 2,000 reviews for each domain from the same review source. We extracted sentences from each review and assigned review's label to its sentences. Like previous work, we treat 4 or 5 stars as positive and 1 or 2 stars as negative. The data is noisy because a positive (negative) review can contain negative (positive) sentences, and there are also neutral sentences. This gives us three noisy-labeled training datasets. We still use the 
same test sets as those for the clean-labeled datasets. Summary statistics of all the datasets are shown in Table \ref{table1}.

\begin{table*}[!htb]
    \centering
    \resizebox{\textwidth}{!}{
    \begin{tabular}{|r|ccc|ccc|ccc|}
    \hline
        Methods & \multicolumn{3}{c|}{\textbf{Movie}} & \multicolumn{3}{c|}{\textbf{Laptop}} & \multicolumn{3}{c|}{\textbf{Restaurant}} \\
    \hline
        & ACC  & F1\_pos & F1\_neg & ACC  & F1\_pos & F1\_neg & ACC  & F1\_pos & F1\_neg \\
        NBSVM-uni \cite{wang2012baselines} & 0.6791 & 0.6663 & 0.6910 & 0.7637 & 0.8216 & 0.6500 & 0.7949 & 0.8478 & 0.6858 \\
        NBSVM-bi \cite{wang2012baselines} & 0.6416 & 0.6438 & 0.6394 & 0.7784 & 0.8320 & \textbf{0.6749} & 0.7154 & 0.7834 & 0.5853 \\
        CNN \cite{kim2014convolutional} & 0.6667 & 0.6467 & 0.6844 & 0.7737 & 0.8381 & 0.6245 & 0.8329 & 0.8841 & 0.7007 \\
        Adaptation \cite{goldberger2017training} & 0.6682 & 0.6708 & 0.6656 & 0.7272 & 0.7936 & 0.5981 & 0.8285 & 0.8872 & 0.6422 \\
        Forward \cite{patrini2017making} & 0.6864 & 0.6753 & 0.6969 & 0.7547 & 0.8170 & 0.6282 & 0.8329 & 0.8882 & 0.6695 \\
        Backward \cite{patrini2017making} & 0.6651 & 0.6160 & 0.6830 & 0.7124 & 0.7834 & 0.5723 & 0.7890 & 0.8485 & 0.6521 \\
        Masking \cite{han2018masking} & 0.6708 & 0.6631 & 0.6782 & 0.7188 & 0.7787 & 0.6144 & 0.8219 & 0.8789 & 0.6639 \\
        Co-teaching \cite{han2018co} & 0.6150 & 0.5980 & 0.6306 & 0.7145 & 0.7867 & 0.5686 & 0.7978 & 0.8575 & 0.6515 \\
    \rowcolor{mygray}\textsc{NetAb} (Our method) & \textbf{0.7047} & \textbf{0.7076} & \textbf{0.7017} & \textbf{0.7928} & \textbf{0.8487} & 0.6711 & \textbf{0.8593} & \textbf{0.9056} & \textbf{0.7241} \\
    \hline
    \end{tabular} }
	\vspace{-0.2cm}
    \caption{Accuracy (ACC) of both classes, F1 (F1\_pos) of positive class and F1 (F1\_neg) of negative class on clean test data/sentences. Training data are real noisy-labeled sentences.}
    \label{table2}
	\vspace{-0.45cm}
\end{table*}

\textbf{Experiment 1:} Here we use the clean-labeled data (i.e., the last three columns in Table \ref{table1}). We corrupt the clean training data by switching the labels of some random instances based on a noise rate parameter. Then we use the corrupted data to train \textsc{NetAb} and CNN \cite{kim2014convolutional}.

\begin{figure}[!htb]
    \centering
    \includegraphics[width=0.48\textwidth]{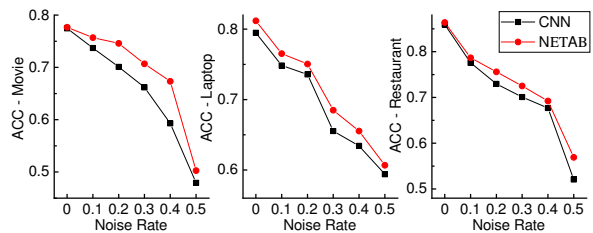}
	\vspace{-0.6cm}
    \caption{Accuracy (ACC) on clean test data. For training, the labels of clean data are flipped with the noise rates [$0$, $0.1$, $0.2$, $0.3$, $0.4$, $0.5$]. For example, $0.1$ means that 10\% of the labels are flipped. (Color online)}
    \label{fig1}
    \vspace{-0.3cm}
\end{figure}

The test accuracy curves with the noise rates [$0$, $0.1$, $0.2$, $0.3$, $0.4$, $0.5$] are shown in Figure \ref{fig1}. From the figure, we can see that the test accuracy drops from around 0.8 to 0.5 when the noise rate increases from 0 to 0.5, but our \textsc{NetAb} outperforms CNN. The results clearly show that the performance of the CNN drops quite a lot with the noise rate increasing.

\textbf{Experiment 2:} Here we use the real noisy-labeled training data to train our model and the baselines, and then test on the test data in Table~\ref{table1}. Our goal is two fold. First, we want to evaluate \textsc{NetAb} using real noisy data. Second, we want to see whether sentences with review level labels can be used to build effective SSC models.


\textbf{Baselines.} We use one strong non-DNN baseline, NBSVM (with unigrams or bigrams features) \cite{wang2012baselines} and six DNN baselines. The first DNN baseline is CNN \cite{kim2014convolutional}, which does not handle noisy labels. The other five were designed to handle noisy labels.

The comparison results are shown in Table \ref{table2}. From the results, we can make the following observations. (1) Our \textsc{NetAb} model achieves the best ACC and F1 on all datasets except for F1 of negative class on Laptop. The results demonstrate the superiority of \textsc{NetAb}. (2) \textsc{NetAb} outperforms the baselines designed for learning with noisy labels. These baselines are inferior to ours as they were tailored for image classification. Note that we found no existing method to deal with noisy labels for SSC. 

\textbf{Training Details.} 
We use the publicly available pre-trained embedding GloVe.840B \cite{pennington2014glove} to initialize the word vectors and the embedding dimension is 300.

For each baseline, we obtain the system from its author and use its default parameters. As the DNN baselines (except CNN) were proposed for image classification, we change the input channels from 3 to 1. For our \textsc{NetAb}, we follow \citet{kim2014convolutional} to use window sizes of 3, 4 and 5 words with 100 feature maps per window size, resulting in 300-dimensional encoding vectors. The input length of sentence is set to 40. The network parameters are updated using the Adam optimizer \cite{kingma2014adam} with a learning rate of 0.001. The learning rate is clipped gradually using a norm of 0.96 in performing the Adam optimization. The dropout rate is 0.5 in the input layer. The number of epochs is 200 and batch size is 50. 

\section{Conclusions}
\vspace{-0.1cm}
This paper proposed a novel CNN
based model for sentence-level sentiment classification learning for data with noisy labels. 
The proposed model learns to handle noisy labels during training by training two networks alternately. The learned noisy transition matrices are used to tackle noisy labels. Experimental results showed that the proposed model outperforms a wide range of baselines markedly. We believe that learning with noisy labels is a promising direction as it is often easy to collect noisy-labeled training data.

\section*{Acknowledgments}
Hao Wang and Yan Yang's work was partially supported by a grant from the National Natural Science Foundation of China (No. 61572407).

\bibliographystyle{acl_natbib}

\end{document}